\documentclass[10pt,journal]{IEEEtran}

\pdfoutput=1

\usepackage{cite}

\ifCLASSINFOpdf
\usepackage[pdftex]{graphicx}
\else
\fi
\usepackage{ifpdf}
\usepackage{times}
\usepackage{epsfig}
\usepackage{graphicx}
\usepackage{amsmath}
\usepackage{bbding}

\usepackage{amssymb}
\usepackage{color}
\usepackage{subfig}
\usepackage{tabu}
\usepackage{tabularx}
\usepackage{multirow}
\usepackage[abs]{overpic}
\usepackage{algorithm}
\usepackage{algorithmic}
\usepackage{wrapfig}
\usepackage{array}
\usepackage{stfloats}
\usepackage{booktabs}
\usepackage[table,xcdraw]{xcolor}
\usepackage{geometry}
\graphicspath{{./Imgs/}{C:/Users/admin/iCloudDrive/paper1/latexpaper1/Imgs/}}    %bu xing zai bu quan lu jing
\DeclareGraphicsExtensions{.pdf,.png,.jpg}
\newcolumntype{C}[1]{>{\centering\arraybackslash}p{#1}}

\newcommand{\figref}[1]{Fig.~\ref{#1}}
\newcommand{\tabref}[1]{Table.~\ref{#1}}

\newcommand{\equref}[1]{Eq. (\ref{#1})}

\def\etal{{\em et al.~}}

\def\sArt{{state-of-the-art~}}

\begin{document}
\title{Salient Instance Segmentation with Region and Box-level Annotations}

\author{Jialun Pei, He Tang, Tianyang Cheng and Chuanbo Chen
  \thanks{* He Tang is the corresponding author.}
  \thanks{Jialun Pei is with the School of Computer Science and Technology, Huazhong University of Science and Technology, 1037 Luoyu Road, Wuhan, 430074, China (e-mail: peijl@hust.edu.cn)}
  \thanks{He Tang, Tianyang Cheng and Chuanbo Chen are with the School of Software Engineering, Huazhong University of Science and Technology, 1037 Luoyu Road, Wuhan, 430074, China (e-mail: hetang@hust.edu.cn; patrickcty@hust.edu.cn; chuanboc@163.com)}
}


\markboth{IEEE TRANSACTION ON CIRCUITS AND SYSTEMS FOR VIDEO TECHNOLOGY,~Vol.~XX, No.~X, XXX~XXXX}%
{Pei \MakeLowercase{\textit{et al.}}: Salient Instance Segmentation with Region and Box-level Annotations}



\maketitle

\begin{abstract}
  Salient instance segmentation is a new challenging task that received widespread attention in the saliency detection area. The new generation of saliency detection provides a strong theoretical and technical basis for video surveillance. Due to the limited scale of the existing dataset and the high mask annotations cost, plenty of supervision source is urgently needed to train a well-performing salient instance model. In this paper, we aim to train a novel salient instance segmentation framework by an inexact supervision without resorting to laborious labeling. To this end, we present a cyclic global context salient instance segmentation network (CGCNet), which is supervised by the combination of salient regions and bounding boxes from the ready-made salient object detection datasets. To locate salient instance more accurately, a global feature refining layer is proposed that dilates the features of the region of interest (ROI) to the global context in a scene. Meanwhile, a labeling updating scheme is embedded in the proposed framework to update the coarse-grained labels for next iteration. Experiment results demonstrate that the proposed end-to-end framework trained by inexact supervised annotations can be competitive to the existing fully supervised salient instance segmentation methods. Without bells and whistles, our proposed method achieves a mask AP of 58.3\% in the test set of Dataset1K that outperforms the mainstream state-of-the-art methods.

\end{abstract}


% \fi
\IEEEpeerreviewmaketitle

\section{Introduction}

\begin{figure*}[!t]
\centering
\includegraphics[width=0.9\linewidth]{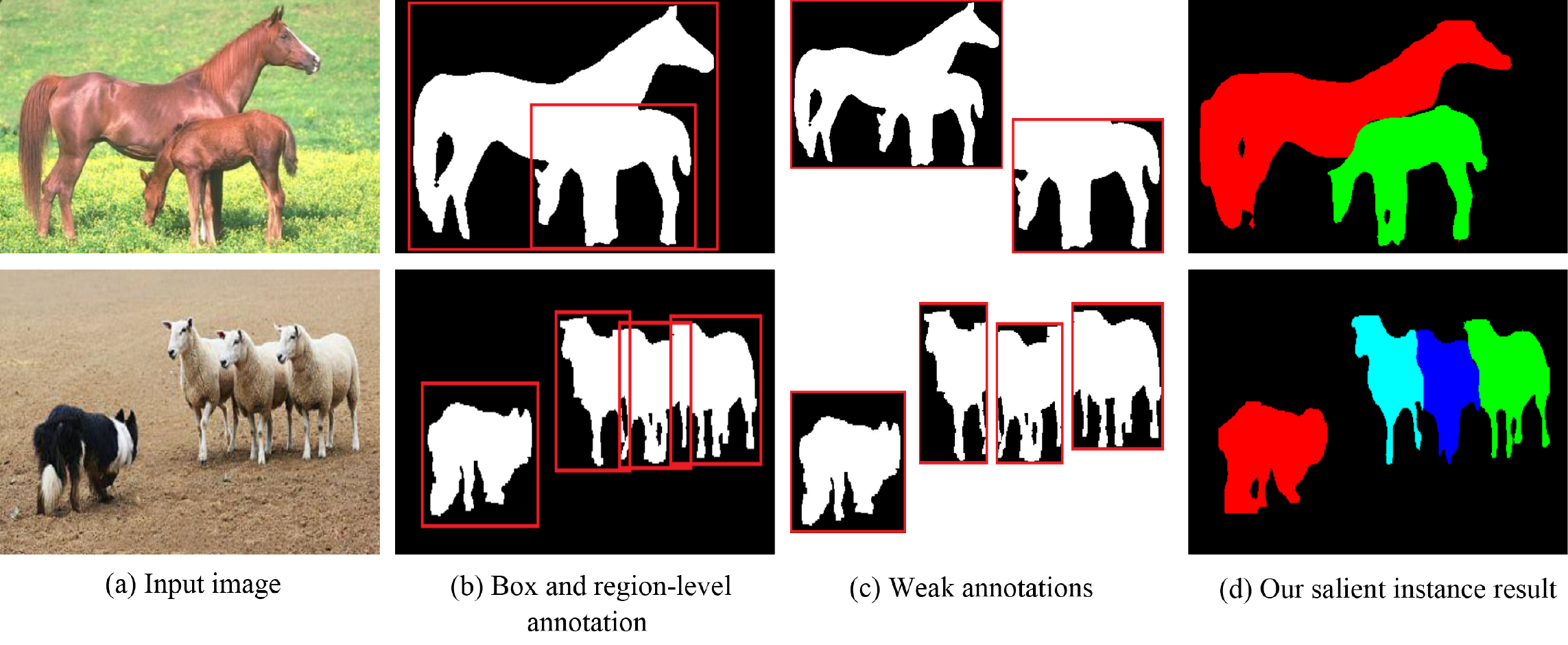}
\caption{The coarse-grained annotation is generated to achieve salient instance results by the proposed framework. (b) shows the combination of bounding box and salient region annotations. (c) exhibits the coarse-grained labels for inexact supervised learning. The final result predicted by CGCNet is showed in (d).}
\label{pipelinemethod}
\vspace{-0.1in}
\end{figure*}

\IEEEPARstart{S}{alient} object detection (SOD) is known as a classic research field for highlighting the most sensitive and informative regions in a scene \cite{han2015background, liu2010learning, zhao2019egnet}. Originating from cognitive and psychology research communities, salient object detection is applied to various areas, such as video surveillance \cite{shao2020surveillance}, video summarization \cite{paul2019spatial} and content-aware image editing \cite{GaoDatabase}. With the rapid development of current image acquisition equipment and 5G communication technology, the traditional binary mask of salient object detection is inadequate to meet the needs of high-resolution image segmentation. Albeit salient object detection task provides the salient region labels compared to the background, they do not explore instance-level cue for salient information. The next generation of salient object detection methods need to showcase more detailed parsing and identify individual instances in salient regions \cite{li2017instance}. In addition, instance-level salient information is more consistent with human perception and offers better image understanding \cite{hsu2019deepco3}. In this paper, we concentrate on the new challenging task salient instance segmentation (SIS) for improving the intelligence level of monitoring systems.

Visual saliency has gained significant progress owing to the rapid development of deep convolutional neural networks (CNNs) \cite{tu2021edge, li2015visual, guo2020motion}. Driven by the strong capability of multi-level feature extraction, CNN models are widely used in the computer vision area \cite{tian2019fcos, pinheiro2016learning}, especially focusing on estimating the bounding boxes of salient instances \cite{zhang2016unconstrained}. Different from salient object detection, salient instance segmentation fosters a more detailed information by labeling each instance with a precise pixel-wise mask and promotes the saliency maps from region-level to instance-level for more detailed analysis. In contrast to instance segmentation, salient instance segmentation only predicts salient instances based on the salient regions. Moreover, segmenting salient instances is class-agnostic compared to the class-specific instance segmentation task.

However, the saliency models of CNNs are usually required to the pixel-level fully-supervised train data \cite{zhu2020aggregate, wang2021deep}. Up to now, the existing SIS dataset is seriously inadequate and the amount of pixel-wise ground-truths is insufficient in a single dataset. The quality and quantity of pixel-level annotations is the bottleneck because the labeling task is strenuous and time-consuming. To alleviate the effectiveness of lacking fully-supervised data, weakly supervised learning is viewed as the alternative training method attracting more attention. This strategy not only avoids user-intensive labeling, but also encourages the models to receive enough training samples.

Inspired by this consideration, in this paper, we aim to integrate the bounding boxes and binary salient regions for training the SIS frameworks. The bounding box annotation contains location information for each salient instance. Meanwhile, salient regions provide salient region information which is a ready-made source generated from the existing SOD datasets. Both box-level and region-level annotations are inexact for salient instance segmentation \cite{zhou2018brief}. As shown in \figref{pipelinemethod}, the bounding boxes determine the location and number of salient instances which have labeled in the DUT-OMRON dataset \cite{yang2013saliency}. We use the bounding box and salient region to assign salient regions to each bounding box of salient instance. It is essential to combine these two supervision sources because the bounding box annotation lacks the pixel-level labels and salient region cannot distinguish different salient instances in the coarse-grained labels. To ensure one instance corresponds to one bounding box and hold the consistency of salient instances and regions, we also exploit some priors to prevent the different object regions trapped into the same box. In this case, the network can utilize more training samples with the lowest labeling cost. We will elaborate the generation steps of the coarse-grained annotations in Section \ref{weaksource}.

\begin{figure*}[!t]
  \centering
  \includegraphics[width=\linewidth]{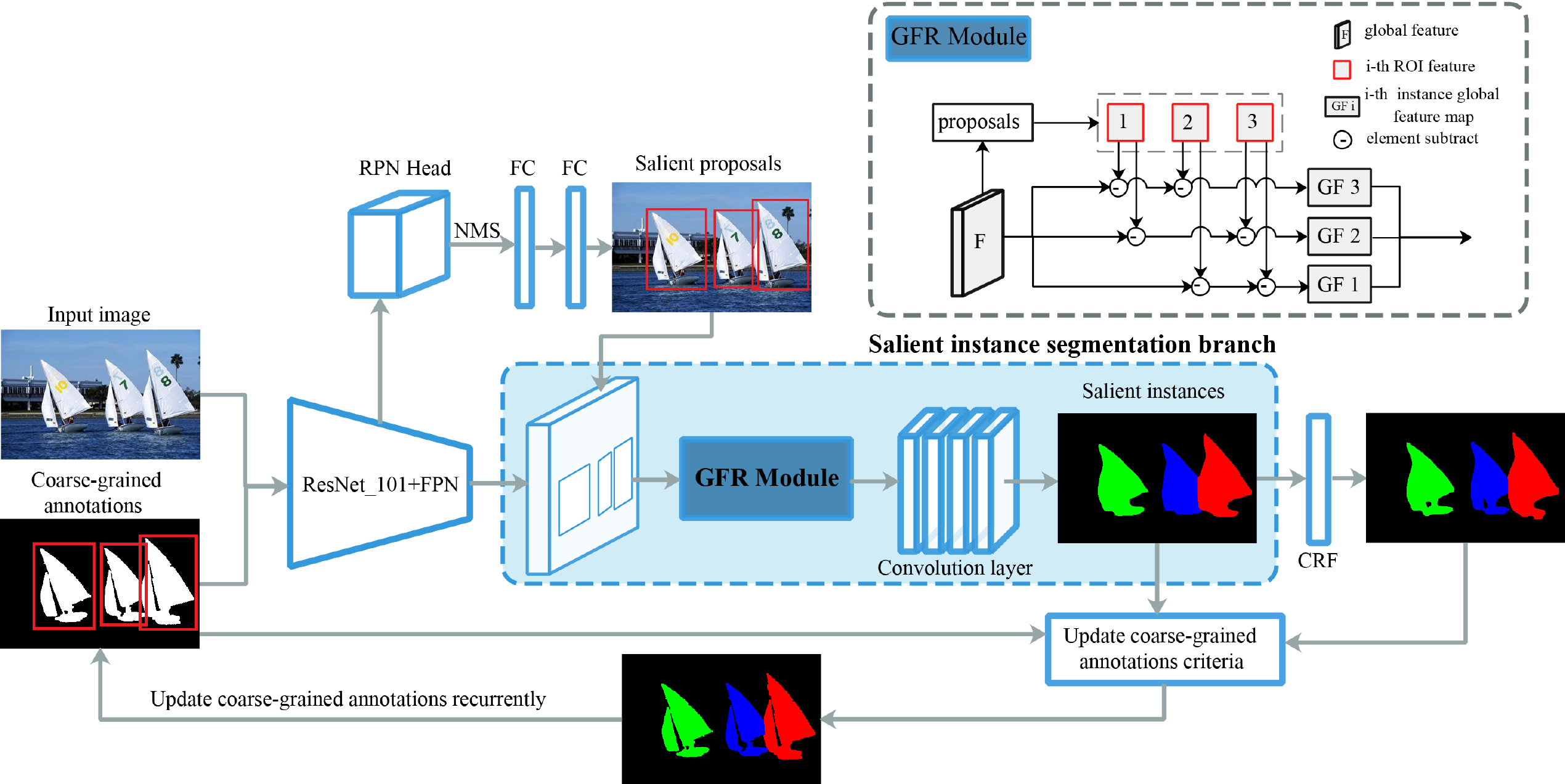}
  \caption{An overview of the proposed framework. The detail of the GFR module is shown in the upper right corner. The coarse-grained annotations updating criteria is illustrated in Section \ref{SISbranch}. At the training time, the salient instance result return to update the coarse-grained annotation in next iteration.}
  \label{figure2}
  \vspace{-0.1in}
\end{figure*}

For segmenting salient instances, we design a cyclic global context SIS network (CGCNet) supervised by the above coarse-grained labels. \figref{figure2} shows the overview of our CGCNet. The proposed model is an end-to-end two-stage SIS framework, which first detects salient proposals and then predict the pixel-level salient instance masks. When extracting features for salient mask prediction, the performance of convolutional layer depends heavily on global context. Considering obtaining stronger feature representation, we extend the scope of feature extraction from the local proposal to the global features. Inspired by enter-surround contrast derived from saliency detection mechanism \cite {klein2011center, xia2016bottom, perazzi2012saliency}, a global feature refining module (GFR) is designed to make full use of background features and suppress disturbance from other salient instance features \cite{long2015fully}. Different from the ROIAlign layer that limits the receptive field in Mask R-CNN \cite{he2017mask}, the proposed GFR module is sensitive to global contrast in order to capture more detailed edge information. Moreover, the CGCNet is designed to iteratively update the coarse-grained annotations by using the forward prediction masks combining with a conditional random field (CRF) \cite{krahenbuhl2011efficient}. It is beneficial to refine the coarse-grained annotations sequentially. The input training samples and the corresponding results are shown in \figref{pipelinemethod}. We evaluate the results on the test set of Dataset1K \cite{li2017instance} and show that our method compares favourably against even some fully supervised methods.

In summary, the main contributions of this paper are as follows:
\begin{itemize}
\item We propose a novel inexact supervision salient instance segmentation framework called cyclic global context network (CGCNet), which is supervised by the combination of the region-level bounding boxes and salient regions.
\item We design a global feature refining (GFR) layer that extends the receptive field of each instance to the global context and suppress the features of other salient instances simultaneously.
\item We embed an update scheme in CGCNet that can optimize the coarse-grained labels continuously to improve the accuracy.
\end{itemize}

The remainder of this paper is organized as follows. Section \uppercase\expandafter{\romannumeral2} presents the related works. Section \uppercase\expandafter{\romannumeral3} describes the architecture and the details of the proposed framework. Section \uppercase\expandafter{\romannumeral4} discusses the experimental settings and comparions with the \sArt methods. Finally, Section \uppercase\expandafter{\romannumeral5} concludes the paper.

\section{RELATED WORK}

\subsection{Salient Object Detection}

Thanks to the fast development of deep learning techniques, salient object detection has gone through a transformation from traditional machine learning to deep CNNs \cite{feng2019attentive}. Driven by the multi-level features extracted from convolution network, the performance of SOD models boost significantly. Fortunately, rich pixel-level salient datasets can be poured into various CNN models to detect salient regions \cite{feng2019attentive, bai2018sod}. Li \etal \cite{li2016deep} proposed a multi-scale deep contrast network to overcome the limitations of overlap and redundancy. Hou \etal \cite{hou2017deeply} designed short connections to the skip-layer structures based on the VGGNet for better supervision. Qin \etal \cite{qin2019basnet} produced a predict-refine SOD network which is composed of a densely supervised encoder-decoder network and a residual refinement module.  Although these SOD methods achieved outstanding performance, the saliency map is viewed as the region-level binary mask which may not accomplish instance-level salient object segmentation.

\subsection{Salient Instance Segmentation}

Proceed from SOD, salient instance segmentation propels the problem into an instance-level phase. Unlike instance segmentation \cite{lee2020centermask, xie2020polarmask, chen2020blendmask}, salient instance is category-independent and it is concentrate on salient regions. Therefore, the frameworks and datasets of instance segmentation are incompatible with segmenting salient instances. Zhang \etal \cite{zhang2016unconstrained} generated salient region-level proposals by CNNs and optimized the bounding boxes based on the Maximum a Posteriori principle. The method is the first to raise saliency detection from the region level to the instance level. Subsequently, Li \etal \cite{li2017instance} formally proposed the instance-level salient object detection task. They drove the prediction results from proposals to pixel-level, and produced the first SIS dataset containing 1,000 samples. Pei \etal \cite{pei2020salient} proposed a multi-task model to predict salient regions and subitizing, and then applied a spectral clustering algorithm to segment salient instances. Recently, Fan \etal \cite{fan2019s4net} proposed an end-to-end single-shot salient instance segmentation framework to segment salient instances. The proposed ROIMasking layer allows more detailed information to be detected accurately, and meanwhile remains the context information around the regions of interest. As a new challenging task, however, the lacking of fully-supervised label is the main problem to limit the performance of deep learning models. To avoid making the high cost of pixel-level annotations, we take advantage of the inexact supervision to train our model.

\subsection{Weakly Supervised Learning}

Most neural networks require full supervision in the form of handcrafted pixel-level masks, which limits their application on large-scale datasets with weaker forms of labeling \cite{zhu2019learning}. To reduce the cost of hand-labelling, weakly supervised learning has attracted a great deal of attention in recent years \cite{bilen2016weakly, cinbis2016weakly, diba2017weakly}. Many weakly supervised principles have been introduced in machine vision area, including object detection, instance segmentation and saliency detection \cite{tang2018weakly, oh2017exploiting}. Weakly supervised learning reveals that the network purposed for one supervision source can resort to another source or incomplete labels. Li \etal \cite{li2018weakly} utilized a coarse activation map from the classification network and saliency maps generated from unsupervised methods as pixel-level annotation to detect salient objects. Zheng \etal \cite{zheng2021weakly} take advantage of salient subitizing as the weak supervision to generate the initial saliency maps, and then propose a saliency updating module (SUM) to refine the saliency maps iteratively. Moreover, Zeng \etal \cite{zeng2019multi} incorporated with diverse supervision sources to train saliency detection models. They designed three networks that learn from category labels, captions and noisy labels, respectively. Inspired by the above contributions, we build an inexact label which embraces the existing binary salient regions and bounding boxes for better training the SIS network.

\section{The CGCNet Architecture}

\subsection{Motivation}

The motivation of the proposed method is handled with segmenting class-agnostic salient instances under lacking fully-supervised annotations. We tend to utilize sufficient training samples with the lowest labeling cost. Therefore, in this paper, the coarse-grained label is proposed that is composed of bounding boxes and binary salient regions. On one hand, the salient proposals provide positional information of salient instance. On the other hand, binary salient regions can provide approximate salient area information for salient instances. Additionally, they can be easily achieved from existing SOD datasets. For training by the coarse-grained labels, we design a cyclic global context neural network (CGCNet) to predict salient instances and update the coarse-grained labels recurrently.

\subsection{Overall Framework}

As shown in \figref{figure2}, the framework of our proposed CGCNet consists of three main components. Firstly, The RPN head is viewed as a salient proposal detector that detects the bounding boxes of salient instance to capture the location and number of salient instances. Then, the GFR module provides the global feature representation to predict salient masks. Moreover, the resulting salient instances update the coarse-grained ground-truth added with the fully connected CRF operation for the next iteration.

We combine pre-trained ResNet-101 \cite{he2016deep} with FPN \cite{lin2017feature} as the backbone. According to the order of downsampling in ResNet-101, we extract the 4-th stage feature map followed by a 1$\times$1 convolutional layer with the lateral connections in multi-level FPN prediction \cite{he2017mask}. Followed by FPN, we utilize five levels of feature maps to detect different sizes of objects on different levels to maximize the gains in accuracy. The feature maps produced by the backbone are extracted from the entire input image. Both salient proposal detector and salient instance segmentation branch are feed with the 256 channel feature maps.

Similar to Faster R-CNN \cite{ren2015faster}, the RPN head is merged into CGCNet for predicting the bounding boxes of each instance in one image. Considering the category-independent characteristic, each ROI feature is assigned to two classes, denoted as $B_c(c\in \left\{ 0,1\right\})$. The two classifications correspond to the background and the salient object in foreground. RPN works on the input features and predicts a pile of salient proposals. Followed by ROIAlign \cite{he2017mask} and two 1024-D Fully Connected layer (FC), the resulting coordinates of salient proposals are generated attached with a confidence score of saliency degree. Then, non-maximum suppression (NMS) \cite{neubeck2006efficient} is embedded to suppress the negative proposals that the saliency score behind the threshold $0.7$ for refining the bounding box of each instance.

The output salient proposals relabel on the feature maps produced by the backbone as input to our GFR module. In this phase, the GFR module extends the ROI feature to the global feature. In addition, this layer retains the feature of the current instance while suppressing the feature of other ROI features. The features processed by the GFR module are injected into a pixel-to-pixel fully convolutional block. The Fully convolutional fashion preserves the spatial consistency of each pixel involved in corresponding salient instances. Moreover, taking the resulting salient instances predicted by the SIS branch, the updating scheme is produced to update the coarse-grained ground-truth recurrently in training phase. In the following subsection, we will describe the SIS branch and the GFR module in detail.

\begin{figure*}[!t]
\centering
\includegraphics[width=\linewidth]{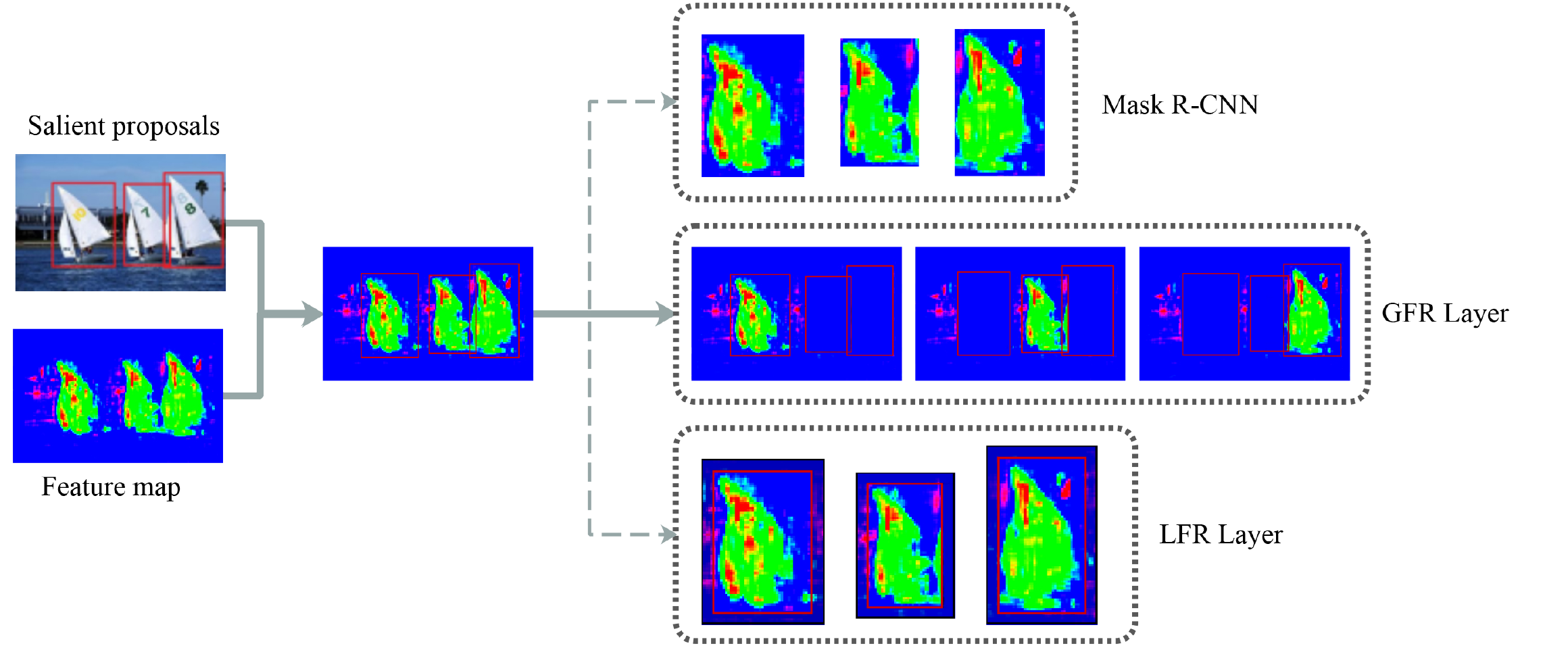}
\caption{Visualization of the GFR module in segmentation branch and comparison of our local feature refining module (LFR module) and Mask R-CNN \cite{he2017mask}.}
\label{figure3}
\end{figure*}

\subsection{Inexact Supervision Sources}\label{weaksource}

We implement the coarse-grained annotations to handle the problem of lacking sufficient labels for the SIS task. Considering the characteristics of salient instances, it is essential to embrace both the salient region and the number of salient instances. Inspired by salient object detection and instance segmentation tasks, the coarse-grained labels are composed of salient regions and the bounding boxes of salient instances. 

To train the proposed CGCNet model, we utilize the largest number of SOD dataset called DUT-OMRON \cite{yang2013saliency}, which contains about 5,000 salient object labels and the bounding boxes. We select 4,500 images from the training set of the DUT-OMRON SOD dataset. Despite combining salient regions and bounding boxes, the coarse-grained labels still have some general issues. First, salient regions from different bounding boxes have shared patches. Second, some small instances are enclosed into the bounding boxes of larger instances. To reduce the negative influence of these obstacles, we provide two priors to deal with ambiguous samples. On one hand, we restrict that each bounding box can contain only one enclosed salient region. On the other hand, if there are multiple closed areas in one bounding box, we only keep the maximal area as its regression target. Given a binary salient map $S$, the bounding box corresponding to each salient instance is denoted as $W_{i} (i=1,2,…,n)$. In addition, we set the patches discarded by priors in each window as $\varphi_{i}$. The final coarse-grained label $I$ is defined by:
\begin{equation}
I=\sum_{i=1}^{n}[S(x,y)\cap W_{i}-\varphi_{i} (\hat{x},\hat{y})],\,\,\,\,i=1,2,...,n,
\end{equation}
where $(x,y)$ presents salient region pixels in the image $S$ and $ (\hat{x},\hat{y})$ denotes the set of pixels excluded by our priors in each window. $n$ is the number of salient instances in an image. The final example can refer to \figref{pipelinemethod}.

\subsection{The Salient Instance Segmentation Branch}\label{SISbranch}

The salient instance segmentation branch aims to segment each salient instance in virtue of the global cues. By achieving the ROI features from the RPN head, we can determine the location and number of salient instances. However, features of each region just contain local spatial information, which is insufficient to segment explicit pixel-level labels. This barrier drives us to explore the broader feature for segmentation. Inspired by center-surround contrast based on the SOD task, we seek to extend the ROI feature to the global feature map. Resorting to increasing receptive field and ensuring the resolution of instances, we utilize global features extracted from the backbone instead of the ROI feature. Meanwhile, each feature map produced from the GFR module only contains the feature of current salient instance proposal and background while suppressing the features of other salient instance proposals.

\textbf{The GFR module.} The goal of the proposed global feature refining module (GFR) is to obtain global context information and limit the disturbance of other instance features.For the ROIAlign module, it only pay attention to the ROI feature and resize the original resolution of ROI \cite{he2017mask}. In S4Net \cite{fan2019s4net}, the ROIMasking extend the receptive field and use of the information around the ROI contrasting ROI features. Differ from the ROIMasking, our GFR module expand each ROI directly to the global feature map and maximize the center-surround contrast for segmenting salient instance.

The internal process in the GFR module is shown in the top right corner of \figref{figure2}. Given the feature maps produced from FPN, the GFR module transfers the coordinates of all proposals from different scales of features to the aspect ratio of original feature map. Tasking $F^{(H\times W \times C)}$ as the input feature map, we assume that the number of proposals is {\em n}. To explain the module more facilitatively, the number of proposals is set to 3. Let $R_{i}^{(H\times W \times C)}(i=1,2,…,n)$ as the feature map includes i-th features of proposals. To maintain the consistency of resolution between {\em F} and $R_{i}$, global average pooling is used to fill in the background area. The output of GFR module $G_{i} (i=1,2,…,n)$ is defined by:
\begin{equation}
G_{i}= F-\sum_{i=1}^{n}R_{i}+R_{i},\,\,\,\,i=1,2,...,n 
\end{equation}
Each feature map $G_{i}$ contains the corresponding feature of proposals and the feature of background. To constrain other features of proposals, the operation of our GFR module first digs out all regions of salient proposals in input feature map and then sticks the corresponding ROI feature on {\em F} according to the coordinates of proposal. This operation also avoids missing the shared pixels from different proposals and reserves the occlusion parts.

\figref{figure3} visualizes the process of GFR module and compares with other similar modules. We also introduce the local feature refining module (LFR). Compared to the GFR module, the LFR module extends the receptive field based on the ROI feature while limiting other salient proposal features rather than covering global features. Assume that the size of salient proposal is ($H_{r}$, $W_{r}$), the size of extended bounding box is set to ($H_{r}+h$, $W_{r}+w$), where $h$ and $w$ is $H_{r}/5$ and $W_{r}/5$, respectively. The other setting of LFR module is same as GFR module. Additionally, the corresponding process in Mask R-CNN \cite{he2017mask} is exhibited in the top branch in \figref{figure3}. The experiment results demonstrate that the GFR module outperforms the other two modules for SIS task, which is discussed in detail in Section \ref{Ablation}.

After adding with our GFR module, each target instance not only contains the features inside the proposal but also takes advantage of the global context information to highlight the instance region. The mask head is efficient to use the contrast of foreground and background features to segment salient instances. For each output feature map from GFR layer, SIS branch stack four consecutive convolutional layers followed on dilated convolutional layer with stride 2 and RELU function \cite{nair2010rectified}. All the convolutional layers have a kernel size 3$\times$3 and stride 1.

\textbf{Coarse-grained Annotations Updating Scheme.} Considering the initial training samples are coarse-grained annotations, we produce an updating scheme to optimize coarse-grained annotations continuously. The fundamental flaw of the coarse-grained labels is that boundary information of each instance is not detailed enough, and different instances in one image have overlap and occlusion. If only training on the original samples, the predicted salient instances would contain some small redundant patches that belong to background or other instances. To further improve the performance of CGCNet, we insert the fully connected conditional random field (CRF) \cite{krahenbuhl2011efficient} after the salient instance maps in the SIS branch because the CRF operation has significant progress on refining the edge of objects. The fully connected CRF model employs the following energy function:
\begin{equation}
E(M)=-\sum_{i}log P(m_i )+\sum_{i,j}\varphi_{p}(m_i,m_j),
\end{equation}
where $M$ presents a binary mask assignment for all pixels, and $P(m_i)$ is the label assignment probability at pixel $i$ belonging to the salient instance. For each binary salient instance mask, the pairwise potential $\varphi_p(m_i,m_j)$ for two labels $m_i$ and $m_j$ is defined by:
\begin{equation}\label{crfcost}
\begin{aligned}
\varphi _{p}\left( m_{i},m_{j}\right) =\omega _{1}\exp \left( -\dfrac {\left| p_{i}-p_{j}\right| ^{2}}{2\theta ^{2}_{\alpha}}-\dfrac {\left| I_{i}-I_{j}\right| ^{2}}{2\theta ^{2}_{\beta }}\right) +\\ w_{2}\exp \left( -\dfrac {\left| p_{i}-p_{j}\right| }{2\theta ^{2}_{\gamma}}\right) ^{2},
\end{aligned}
\end{equation}
where the first kernel depends on pixel positions $p$ and pixel intensities $I$. The kernel encourages nearby pixels with similar features to take consistent salient instance labels \cite{li2016deep}. The second kernel quantifies the smoothness kernel which only depends on pixel positions for removing small isolated regions \cite{ShottonTextonBoost}. $\omega_{1}$ and $\omega_{2}$ indicate the weighted values to balance the two parts. The hyper parameters $\theta_{\alpha}$, $\theta_{\beta}$ and $\theta_{\gamma}$ control the degree of the Gaussian kernels. In this paper, we adopt the publicly available implementation of \cite{krahenbuhl2011efficient} to optimize these parameters. Specifically, we cross-validate the hyperparameters $\omega_{1}$, $\omega_{2}$, $\theta_{\alpha}$, $\theta_{\beta}$ and $\theta_{\gamma}$ for the best performance of CRF. The coarse-to-fine scheme is applied on the subset of validation set (about 100 images) in DUT-ORMON dataset. The default value of $\omega_{2}$ and $\theta_{\gamma}$ are set to 3 and 1, and the initial search range of the parameters are $\omega_{1}\in[1$:$1$:$10]$, $\theta_{\alpha}\in[50$:$5$:$100]$ and $\theta_{\beta}\in[5$:$1$:$15]$. These parameters are fixed through 10 iterations of the average field to achieve the best value. In our experiments, the values of $\omega_{1}$, $\omega_{2}$, $\theta_{\alpha}$, $\theta_{\beta}$ and $\theta_{\gamma}$ are set to 4, 3, 70, 13, 1, respectively.

We denote the salient instance map as {\em R} and the map processed by CRF as $R_{f}$. The coarse-grained annotation is labeled as {\em C}. According to Algorithm 1, we propose a strategy based on the KL-Divergence \cite{cornia2018predicting} to update the {\em C} for the next iteration. KL-Divergence is defined as a dissimilarity metric and a lower value indicates a better approximation between the predicting salient instance maps and the ground-truth. Due to ground-truth of CGCNet is noisy, the updating prediction map should have more dissimilar patches with coarse-grained annotation as well as the larger value of KL-Divergence between them. Our strategy compares the prediction map {\em R} and $R_{f}$ to the coarse-grained annotation {\em C}, which is designed as:
\begin{equation}\label{equk1}
K_{1}(R,C)=\frac{1}{H\times W}\sum_{i=1}^{H\times W}C_ilog(\frac{C_i}{R_i+\sigma }+\sigma )
\end{equation}
\begin{equation}\label{equk2}
K_{2}(R_f,C)=\frac{1}{H\times W}\sum_{i=1}^{H\times W}C_ilog(\frac{C_i}{{R_f}_i+\sigma }+\sigma ), 
\end{equation}
where $K_{1}$ and $K_{2}$ denote the mean KL-Divergence value of {\em R} and $R_{f}$ to {\em C}, respectively. The index of {\em i} is set as the {\em i-th} pixel and $\sigma$ is a regularization constant. In Algorithm 1, $C_{n}$ represents the ground-truth to be used for the next iteration. It is observed that we set $\varphi$ as the threshold to determine whether to update with the existing coarse-grained annotation $C$. The value of $\varphi$ is set to 0.05. The strategy can eliminate redundant replacements and alleviate the impact of excessive erosion of the CRF on the prediction map. Using the updating scheme to the inexact supervised learning, the network achieved more accurate results at the training phase.

\begin{algorithm}
\renewcommand{\algorithmicrequire}{\textbf{Input:}}
\renewcommand{\algorithmicensure}{\textbf{Ensure:}}
\caption{Coarse-grained annotations updating}
\label{alg:1}
\begin{algorithmic}[1]
\REQUIRE Coarse-grained annotation {\em C}, salient instance map {\em R} and salient instance map with CRF $R_{f}$.
\ENSURE The updated coarse-grained annotation $C_{n}$
\STATE \textbf{if} $K_{2}(R_f,C)-K_{1}(R,C)\geq \varphi$
\STATE \textbf{then} $C_{n}=C$
\STATE \textbf{else} $C_{n}=R_{f}$
\STATE \textbf{end if}
\end{algorithmic}
\end{algorithm}

\textbf{Loss Function.} The proposed CGCNet need to trained salient proposal branch and SIS branch simultaneously. Therefore, we use ground-truth proposals to supervise the RPN head and the pixel-level coarse-grained labels to train SIS branch. The loss function of CGCNet is defined as a two-stage fashion: 
\begin{equation}
L= L_{bb}+L_{seg}+L_{upd} 
\end{equation}
Where the $L_{bb}$ function includes a classification loss which is log loss over two classes including saliency or background and a bounding box loss which is similar with $L_{loc}$ in Fast R-CNN \cite{girshick2015fast}. The SIS branch loss $L_{seg}$ is defined by the cross-entropy loss, which is followed by:
\begin{equation}
L_{seg}=-\frac{1}{N}\sum_{i=1}^{N}(g_{i}logp_{i}+(1-g_{i})log(1-p_{i}))
\end{equation}
where $p_{i}$ denotes the probability of pixel $i$ belonging to class $c={0,1}$, and $g_{i}$ indicates the ground truth label for pixel $i$. Inspired by the updating criterion from \equref{equk1} and (\equref{equk2}, the loss function $L_{upd}$ for updating SIS branch for pixel-level salient instance prediction is:
\begin{equation}
L_{upd}=K_{2}(R_f,C)-K_{1}(R,C) 
\end{equation}
In the training phase, the weights of the backbone are frozen. The entire procedure is repeated iteratively for training.

\section{Experimental Results}

In this section, we elaborate on the results of the proposed CGCNet framework for the SIS task in detail. We perform ablation experiments on various components of our approach. Besides, we use different metrics to compare with the experimental results of other \sArt methods. Since the proposed method accomplishes the SIS task by inexact supervised learning, we will maintain maximum fairness in comparison.

\subsection{Implementation Details}

As described in the section above, the end-to-end CGCNet is trained by our inexact labels which select 4,500 images from DUT-OMRON dataset \cite{yang2013saliency} without ambiguous samples. During training, the salient bounding box ground-truths are used to supervise the salient proposal detector while combining with SOD annotations to train the mask branch. Meanwhile, we utilize 500 images as the same as training data for validation. For training salient proposals, the bounding boxes are considered as a positive sample if the IOU is more than 0.7 or a negative sample below 0.3. In addition, the NMS threshold used on the proposal detector is set to 0.7. At inference time, we only use 300 images from the testing set in the dataset proposed in \cite{li2017instance} due to shortage of datasets. We input the number of top 80 scoring proposals from the proposal prediction branch after applying NMS to the GFR module. Additionally, the SIS branch directly outputs the resulting images without the updating scheme.

Our proposed framework is implemented in PyTorch framework on 2 NVIDIA GeForce GTX 1080Ti GPUs with 22 GB of memory. To speed up training convergence, we initialize the CGCNet with a pre-trained model over the ImageNet dataset \cite{deng2009imagenet} from Mask R-CNN \cite{he2017mask}. The CGCNet is fine-tuned by flipping the training sets horizontally at a probability of 0.5. In our experiments, we train our network with a learning rate of 0.0025 which is decreased by 10 at the 8K iteration. The training process totally iterates 16K times by using the batch size of 4. The weight decay is empirically set to 0.0001 and the momentum is 0.9.

\begin{table}[!t]
    \centering
    \scriptsize
    \renewcommand{\arraystretch}{1.5}
    \renewcommand{\tabcolsep}{6mm}
    \caption{Comparison of different backbones used in the CGCNet on DUT-ORMON validation set. In this experiment, we keep the rest part of the framework in line.}\label{Table1}
    \begin{tabular}{cccc}
    \toprule
    Backbone                 & AP    & AP$^{r}$0.5 & AP$^{r}$0.7 \\ \hline
    VGG16 \cite{simonyan2014very}           & 50.79 & 79.28  & 60.38  \\ 
    ResNet-50 \cite{li2018weakly}       & 57.13 & 85.6  & 71.02 \\
    ResNet-101 \cite{wang2017learning}      & 57.69 & 86.04  & 71.72  \\ 
    ResNeXt-101 \cite{xie2017aggregated} & \textbf{58.28} & \textbf{86.91}  & \textbf{72.69}  \\ 
    \bottomrule
    \end{tabular}
\end{table}

\subsection{Evaluation Metrics}

For a brand new task, salient instance segmentation has few evaluation metrics to measure its performance quantitatively. Different from SOD and instance segmentation, The SIS task distinguishes pixel-level instances based on salient regions without classification. Therefore, we adopt the $AP$ metric to calculate the average of maximum precision value at IoU scores of 0.5 and 0.7 instead of MAP metric \cite{hariharan2014simultaneous}. The precision value of one image is computed by the predicted number of salient instances (IoU $>$0.5 or 0.7) divided by the real number of salient instances in the image. So, the $AP^{r}$ metric is defined by the summation of precision value divided by the number of all images in testing set, which is formulated as:
\begin{equation}
AP^{r}\alpha =\dfrac {1}{N}\sum _{j}\dfrac {1}{n}\sum _{i}precision,\,\,\,\,IoU(i)\geq \alpha 
\end{equation}
\begin{equation}
precision=\begin{cases}1,\,\,\,\,if\,\,IoU(i)\geq \alpha \\ 0,\,\,\,\,if\,\,IoU(i)< \alpha \end{cases},
\end{equation}
where $\alpha$ is the threshold of IoU. $N$ is the number of instances in one image and $n$ is the total instances in the dataset. Moreover, the $AP$ metric is used to measure the effectiveness of salient instance segmentation according to the $AP^{r}$ metric. The metric average the $AP^{r}$ metric with the threshold of IoU from 0.5 to 0.95 by step 0.05, which is calculated by :
\begin{equation}
AP=\frac{1}{10}\sum _{\alpha }AP^{r}|\alpha ,\,\,\,\,\alpha=0.5,0.55,...,0.95 
\end{equation}
Compared with the $AP^{r}$ metric, the $AP$ value is adopted to measure the overall performance of SIS methods. In this section, the experimental results are evaluated mainly based on the above-mentioned two metrics.

\begin{table}[!t]
    \centering
    \scriptsize
    \renewcommand{\arraystretch}{1.5}
    \renewcommand{\tabcolsep}{2mm}
    \caption{Ablation study for different modules in SIS branch. The experiment is evaluated on DUT-ORMON validation set.}\label{Table2}
    \begin{tabular}{ccccc}
    \toprule
    Modules & LFR module & GFR module & ROIAlign \cite{he2017mask} & ROIMasking \cite{fan2019s4net} \\ \hline
    AP$^{r}$0.5  & 85.45     & \textbf{86.04}                                        & 85.25    & 85.73      \\
    AP$^{r}$0.7  & 70.2     & \textbf{71.72}                                        & 70.28    & 70.46     \\ 
    \bottomrule
    \end{tabular}
\end{table}

\subsection{Ablation Studies}\label{Ablation}

We analyze the effectiveness of the proposed CGCNet on DUT-OMRON validation set \cite{yang2013saliency}. The ablation studies contain four parts: performance of four different backbones, performance of GFR module versus three related structures, hyper-parameter of the updating scheme and contributions of each component of our framework.

\textbf{Backbone:} To ensure fairness and the effects of the different backbones on the experimental results, we verify various backbones working on CGCNet which stay in the same settings. \tabref{Table1} shows the effectiveness of these base models working on the framework. It demonstrates that the backbone of combining ResNeXt-101 achieves the best performance whether $AP$ or $AP^{r}$ metric \cite{xie2017aggregated}. The widely used ResNet-101 has also achieved good results slightly behind ResNeXt-101. Due to insufficient depth of the network, VGGNet obtained relatively low accuracy, but is slightly faster than ResNet \cite{simonyan2014very}.

\textbf{The GFR Module:} The proposed GFR module is viewed as the core layer in SIS branch to refine features. In this section, we try to evaluate the feature refining layer containing local and global cues, respectively. \tabref{Table2} lists the performance of the LFR module and GFR module. Meanwhile, we also compare similar methods embedded in the segmentation branch based on CGCNet, including ROIAlign in Mask R-CNN \cite{he2017mask} and ROIMasking in S4Net \cite{fan2019s4net}. As shown in \tabref{Table2}, the experimental results based on GFR module outperforms other modules. ROIAlign only concentrates on the ROI features. Albeit the LFR module extended the scale of features around ROI, it still slightly behind the ROIMasking by reason of its ternary masking. It indicates that treatment of refining features play an important role in segmenting salient instances. Finally, we adopt the GFR module embedded in our framework.

\textbf{Hyper-parameter in updating scheme:} The threshold $\varphi$ of updating scheme is essential for the quality of inexact supervised annotations to train our framework. In our experiment, we find the appropriate threshold to ensure the efficiency at the training time. According to the formulation of KL-Divergence \cite{krahenbuhl2011efficient}, we empirically provide several default values for determining its influence in this experiment, which is shown in \tabref{Table3}. The performance of different values of $\varphi$ is relatively average. The best result is obtained when the value of $\varphi$ was set to 0.05, it can balance the optimal quantity and quality of replacement.

\begin{figure*}[!t]
\centering
\includegraphics[width=0.8\linewidth]{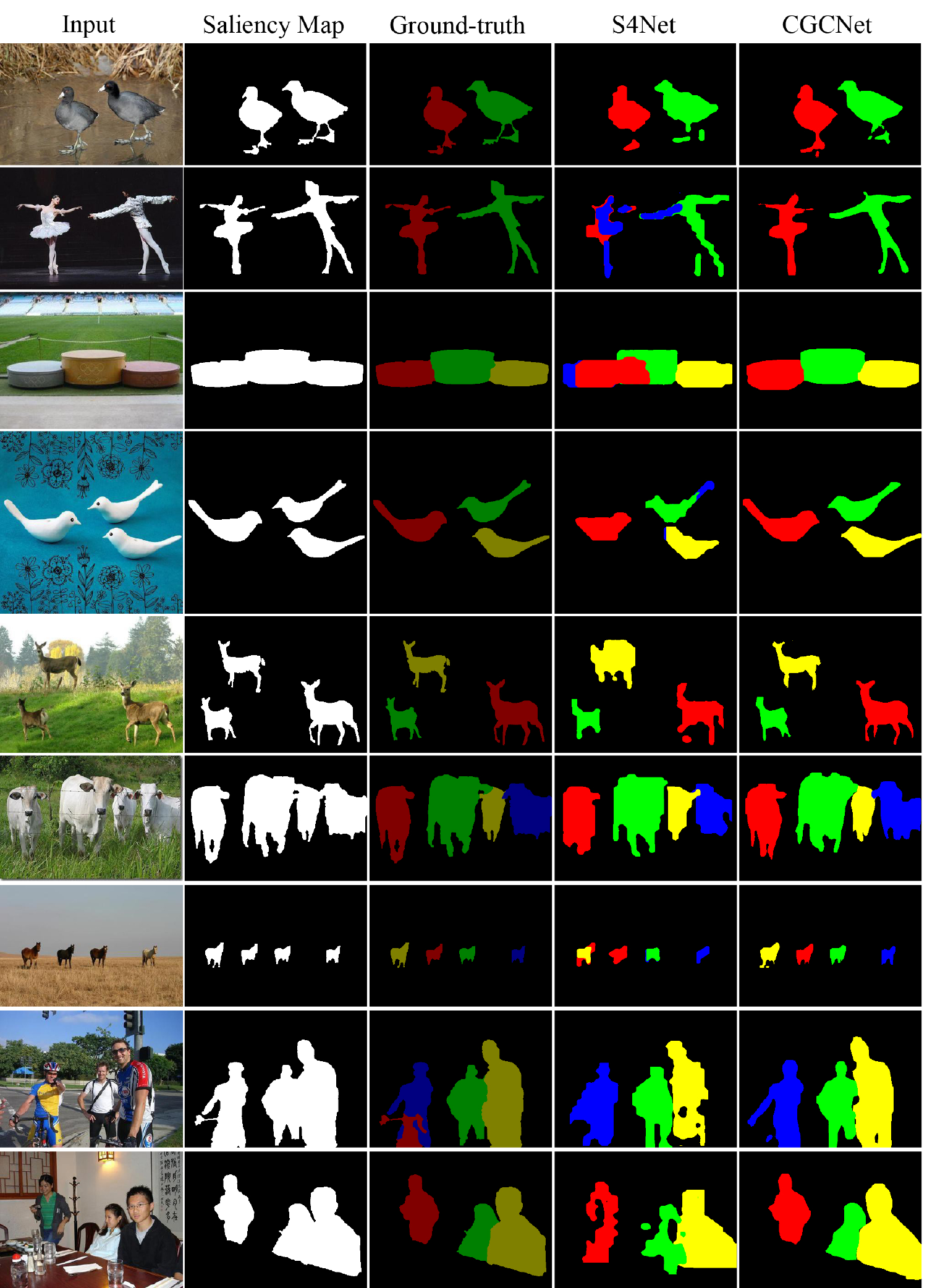}
\caption{Qualitative analysis of experimental results by the proposed method and S4Net \cite{fan2019s4net}.}
\label{figure4}
\end{figure*}

\begin{table}[!t]
    \centering
    \scriptsize
    \renewcommand{\arraystretch}{1.5}
    \renewcommand{\tabcolsep}{4mm}
    \caption{The threshold $\varphi$ of updating scheme performance of CGCNet. The highest scores in each row are labeled in bold.}\label{Table3}
    \begin{tabular}{cccccc}
    \toprule
    $\varphi$ & 0.01  & 0.05  & 0.1   & 0.15  & 0.2   \\ \hline
    AP$^{r}$0.5                 & 85.89 & \textbf{86.04} & 84.85 & 84.66 & 84.13 \\
    AP$^{r}$0.7                 & 71.34 & \textbf{71.72} & 71.16 & 70.68 & 70.19  \\
    \bottomrule
    \end{tabular}
\end{table}

\textbf{The component in CGCNet:} We conducted extensive experiments to discover contributions of each innovative module under the same settings. These parts of CGCNet include the prior criteria (Standardized coarse-grained labels), the updating scheme and the GFR module. As shown in \tabref{Table4}, the various parts of our framework have various degrees of contribution for segmenting salient instances. Particularly, the updating scheme has more contributions that improved the $AP$ metric about 2 percent compared to without it. It can be attributed to the insertion of CRF and the revision of the coarse-grained annotations at the training time. With the help of the prior criteria, the performance significantly improved in terms of $AP^{r}$0.5 and $AP^{r}$0.7 metrics. Overall, each module has an indispensable contribution to the entire framework.

\begin{table}[!t]
    \centering
    \scriptsize
    \renewcommand{\arraystretch}{1.3}
    \renewcommand{\tabcolsep}{3mm}
    \caption{Ablation analysis of effects of various components from our model on SIS task. PC, GFR and US means the prior criteria, the GFR module and the updating scheme, respectively. The experiment is evaluated on DUT-ORMON validation set.}\label{Table4}
    \begin{tabular}{llll}
    \toprule
    Models    & AP & AP$^{r}$0.5    & AP$^{r}$0.7    \\ \hline
    The basic model & 53.93     & 83.44     & 66.86     \\
    The basic model + PC    & 54.67  & 85.81     & 68.43   \\
    The basic model + PC + GFR  & 55.84  & 85.15  & 70.54   \\
    The basic model + PC + GFR + US  & \textbf{57.69} & \textbf{86.04} & \textbf{71.72}   \\
    \bottomrule
    \end{tabular}
\end{table}

\begin{table}[!t]
    \centering
    \scriptsize
    \renewcommand{\arraystretch}{2.0}
    \renewcommand{\tabcolsep}{2.9mm}
    \caption{Quantitative comparisons with existing methods on the training set of our inexact labels and Dataset1K \cite{li2017instance}, respectively. The results are evaluated on the test set of Dataset1K \cite{li2017instance}. For a fair comparison, both our method and S4Net \cite{fan2019s4net} use ResNet-50 as backbone. We keep the rest part of the framework in line. '-' indicates unacquirable value.}\label{Table5}
\begin{tabular}{c|c|c|c|c}
\hline
Method        & Training Set                    & AP             & AP$^{r}$0.5    & AP$^{r}$0.7    \\ \hline
S4Net \cite{fan2019s4net}        &   \multirow{2}{*}{\begin{tabular}[c]{@{}c@{}}DUT-ORMON\\ (Inexact labels)\end{tabular}}     & 50.9          & 84.9          & 60.8         \\ \cline{1-1} \cline{3-5} 
CGCNet (Ours) &                                 & \textbf{58.3} & \textbf{88.4} & \textbf{71.0} \\ \hline
MSRNet \cite{li2017instance}     & \multirow{4}{*}{Dataset1K \cite{li2017instance}}         & -              & 65.3          & 52.3          \\ \cline{1-1} \cline{3-5} 
SCNet \cite{pei2020salient}      &                                 & 56.8           & 84.6           & 67.4           \\ \cline{1-1} \cline{3-5} 
S4Net \cite{fan2019s4net}        &                                 & 52.3           & \textbf{86.7}  & 63.6           \\ \cline{1-1} \cline{3-5} 
CGCNet (Ours)         &                                 & \textbf{57.1} & 85.8           & \textbf{69.0} \\ 
\hline
\end{tabular}
\end{table}

\begin{figure*}[!t]
    \centering
    \includegraphics[width=\linewidth]{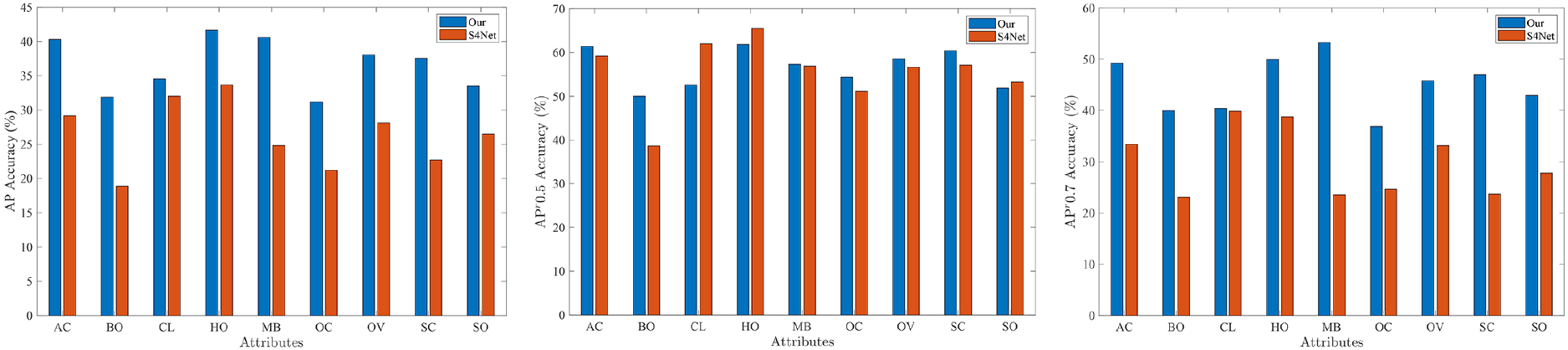}
    \caption{The attributes-based performance of the CGCNet on the instance-level SOC test set. The left of histogram shows the accuracy of $AP$ metric. The histograms in the middle and right show the accuracy of $AP^{r}$0.5 and $AP^{r}$0.7 metric under nine attributes.}
    \label{figure5}
    \vspace{-0.1in}
\end{figure*}

\begin{figure*}[!t]
    \centering
    \includegraphics[width=0.8\linewidth]{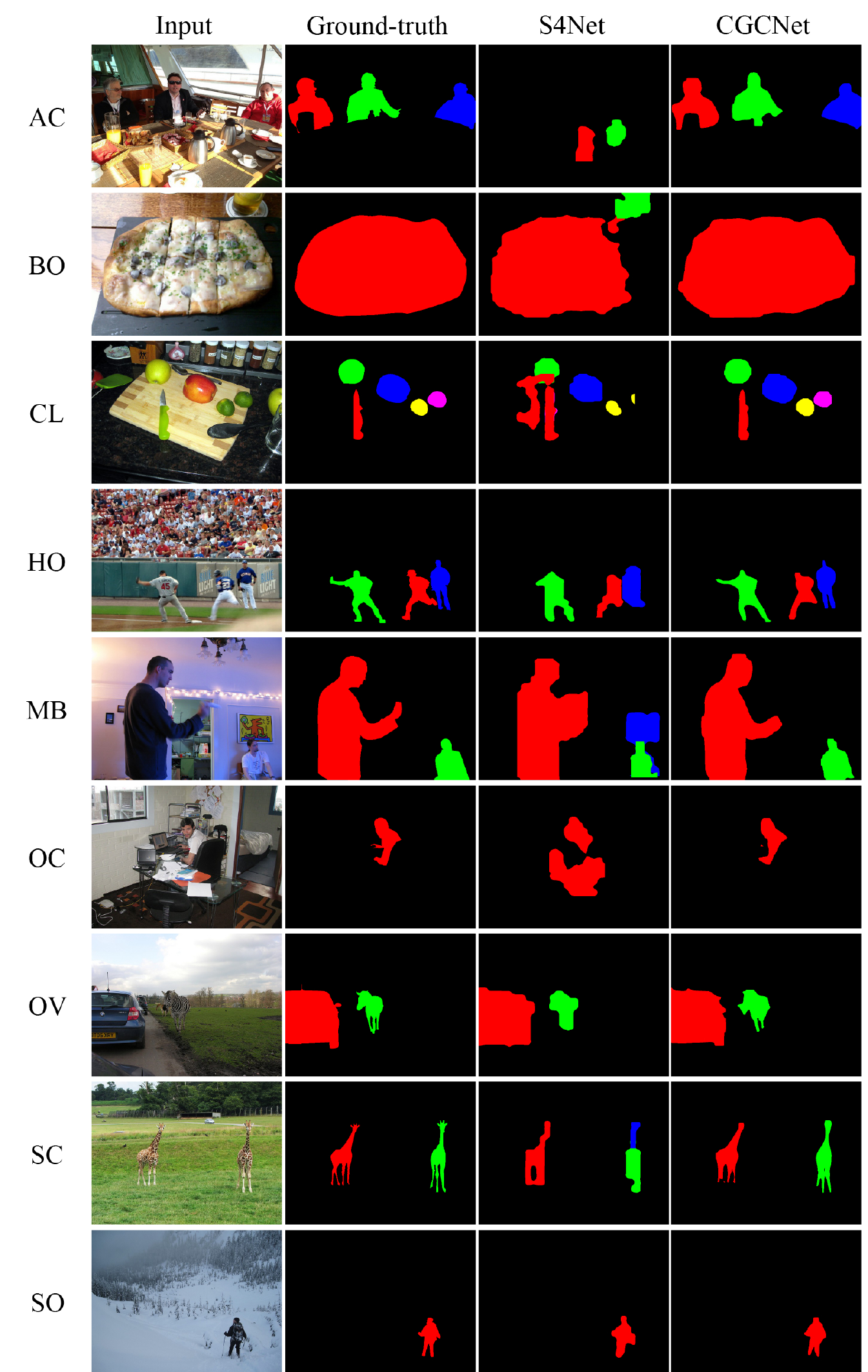}
    \caption{Representative experimental results for each attribute produced by S4Net and the proposed method. Both frameworks are fine-tuned on the Dataset1k training set \cite{li2017instance} and tested on the SOC test set \cite{fan2018SOC}. We select a most representative sample in each attribute-based test subset. Each row displays one attribute. We keep the setting of two framework in line.}
    \label{figure6}
\end{figure*}

\begin{figure}[!t]
    \centering
    \includegraphics[width=\linewidth]{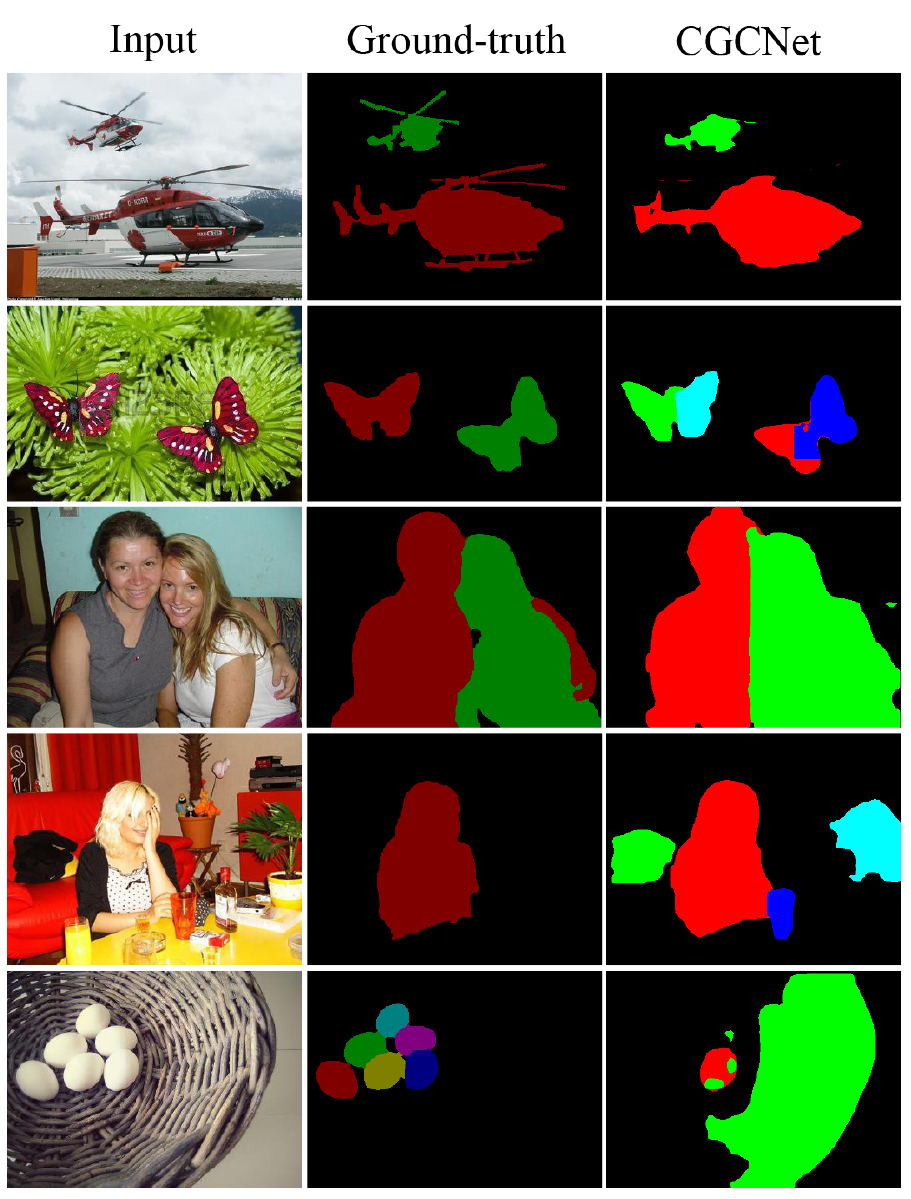}
    \caption{Example of failure modes generated by our method. Samples are selected from the Dataset1k test set \cite{li2017instance}}
    \label{failureexamples}
    \vspace{-0.1in}
\end{figure}

\subsection{Comparison with the \sArt Methods}

There are three existing methods related to the salient instance segmentation task: MSRNet \cite{li2017instance}, S4Net \cite{fan2019s4net} and SCNet \cite{pei2020salient}. In contrast to these previous works, we are the first to make use of inexact supervised learning for the new challenging task. All methods are evaluated on the test set of Dataset1K \cite{li2017instance} and SOC dataset \cite{fan2018SOC}, respectively. For fair comparison, we compare the existing salient instance segmentation methods qualitatively and quantitatively on the only two datasets.

\textbf{Evaluation on the Dataset1K:} The Dataset1K \cite{li2017instance} is the first salient instance dataset, which contains 500 images for training, 200 images for validation and 300 images for testing. Considering that all existing methods are fully supervised and our method is supervised by inexact labels, we train all methods on the training set of Dataset1K and our coarse-grained annotations of DUT-OMRON dataset, respectively. Then, we evaluate these models on the test set of Dataset1K \cite{li2017instance}. Since our inexact labels are not applicable to MSRNet and SCNet, we only compare with S4Net by using inexact labels for training. The proposed CGCNet use ResNet-50 as backbone to stay the same with S4Net. Other settings also maintain relative consistency and fairness in this experiment. \tabref{Table5} lists the value of $AP$, $AP^{r}$0.5 and $AP^{r}$0.7 metric achieved by different training set. Due to the related code of \cite{li2017instance} is not available, we cannot obtain its whole results. In the case of training on the inexact labels, our method achieves the best result compared to all other methods. As an inexact supervised method, the CGCNet improves the value of $AP$ metric to the highest 58.3\%. Additionally, we also exhibit the results of our framework and other fully-supervised methods on the training set of Dataset1K \cite{li2017instance}. As shown in the bottom of \tabref{Table5}, the value of $AP$ achieved by our CGCNet also outperforms SCNet and S4Net. While the value of $AP^{r}$0.5 metric is slightly lower than S4Net, our framework has demonstrated its robustness whether trained on inexact labels or not.

We also qualitatively analyzed the experimental results produced by CGCNet and S4Net. \figref{figure4} displays some results from the testing set in Dataset1K \cite{li2017instance}. It shows that our method produces high quality results which is very close to the ground-truth. The first two input images contain two instances, which have similar internal features and relatively simple backgrounds. Our method can easily segment salient instances from the background. The middle images in \figref{figure4} have multiple instances and each instance is close together. Our model can still predict the number of instances accurately and segments them effectively. The last two samples have chaotic backgrounds, and the internal features of salient instances are also very messy. In this complex case, the CGCNet also distinguish obstructed instances satisfactorily. In comparison, the S4Net determine the number of salient instances inaccurately in some cases. The antepenult sample demonstrated that the S4Net is insensitive to smaller salient instances. In addition, our method is better than S4Net in smoothing the edge of salient instances. It indicates that the lack of fully supervised data limits the performance of S4Net. By and large, the proposed framework has high accuracy and robustness for salient instance segmentation.

\textbf{Evaluation on the SOC}: Recently, Fan \etal \cite{fan2018SOC} introduced a Salient Object in Clutter dataset called SOC, which contains both binary mask and instance-level salient ground-truth. Considering that the dataset labels salient instances in clutter, the difficulty of input images is relatively high. Therefore, the experiment results will be lower than other datasets. In this experiment, we analyze the proposed CGCNet in terms of image attributes on the test set of SOC dataset. The instance-level test set is divided into nine attributes: Appearance Change (AC), Big Object (BO), Clutter (CL), Heterogeneous Object (HO), Motion Blur (MB), Occlusion (OC), Out-of-View (OV), Shape Complexity (SC) and Small Object (SO) \cite{fan2018SOC}. We compare the experimental results of S4Net according to the attributes. For fair comparison, both methods are trained on Dataset1K training set \cite{li2017instance}, and then directly tested on the SOC test set. The histograms in \figref{figure5} show the performance of the CGCNet and S4Net on different attribute test subsets. Although these two methods achieve approximate scores in terms of $AP^{r}$0.5 metric, CGCNet performs significantly in $AP$ values. It can be attributed to the better suppression of complex background by the GFR module. The right histogram demonstrates that the proposed method is more generalized for images with different attributes. Moreover, our framework excels at dealing with the image containing heterogeneous object (HO) compared to other attributes. Thanks to the global features of the GFR module, CGCNet process the image with AC attribute effectively. The $AP$ value of OC attribute is lowest because the occluded part of object is difficult to detect. Overall, our method is robust for processing images with different attributes.

\figref{figure6} exhibits some typical results generated by S4Net and our framework according to different attributes. Compared to the Dataset1K, the test set in SOC contains more different kinds of images and the complexity of background is higher. Our method also shows great performance on the SOC dataset against the S4Net. For example, the sample in the first row has the obvious illumination change in salient instance area combining with messy background, the proposed method can easily dig out salient instances from background. The Clutter-based (CL) image has several small salient instances, and the foreground and background regions around instances have similar color. The proposed CGCNet can still accurately locate each instance and segment them out. Refer to the last two rows of \figref{figure6}, salient instances in images with SC and SO attributes have complex boundaries and are relatively small. Although it is not easy to split the slender legs of the giraffe, the overall result is satisfying.

\textbf{Limitations}: \figref{failureexamples} displays some typical failure cases. According to the first row, our method is insensitive to the tenuous local features. Due to the two-stage framework, it is inefficient to suppress the number of proposals in the second row. This strategy tends to result in a greater number of predicted salient instances than the ground-truth. The third row shows that the detail of the boundary is terrible when two salient instances overlap. It is due to the inexact annotations consisting of bounding boxes and salient regions that cause the edge of the salient instance to become the edge of boxes. The bottom two cases demonstrate that our approach fails to predict the salient regions. The problem is very common in saliency detection tasks. Generally, it is beneficial to use coarse-grained labels based on the proposed CGCNet.

%---------------------------------------------------------------
\section{Conclusion}

In this paper, we propose an end-to-end cyclic global context neural network (CGCNet) for salient instance segmentation. Due to lack of dataset for the new challenging task, we used inexact supervised learning to train our framework. More importantly, adding with the GFR module and the updating scheme in CGCNet, our framework shows excellent performance for salient instance segmentation, which compares favorably against even some fully supervised methods. Due to dependence on the post processing of NMS, the framework sometimes predicts the number of salient instances inaccurately. In the future work, we will attempt to exploit one-stage single network and further improve the effectiveness of the framework for applying to video surveillance.

%\begin{figure}[!t]
%    \centering
%    \includegraphics[width=0.6\linewidth]{Figure5}
%    \caption{Confusion matrix of the proposed DSN. Each cell includes the accuracy percentage (recall). %Each row corresponds to the ground-truth number, while each column shows the predicted results. The %values on the diagonal line represent the correct result percentages.}
%    \label{Confusion matrixDSN}
%    \vspace{-0.1in}
%\end{figure}

%\begin{table}[!t]
%    \renewcommand{\arraystretch}{1.3}
%    \caption{Comparison of the DSN and different versions of DenseNet model. The second row represents 
%5the average precision $(\%)$ of the predicted number of instances.}
%    \label{tab:Densenet}
%    \centering
%    \resizebox{0.48\textwidth}{!}{
%\begin{tabular}{c|c|c|c|c}
%\hline
%DSN  & DenseNet-121 & DenseNet-161 & DenseNet-169 & DenseNet-201 \\ \hline
%79.4 & 68.7         & 72.5         & 70.3         & 77.5         \\ \hline
%\end{tabular}}
%\end{table}

% or

\section*{Acknowledgments}
This research was supported by the National Natural Science Foundation of China Grant 61902139.

%\appendices

%\section{}

%\section*{Acknowledgment}


%\ifCLASSOPTIONcaptionsoff
%  \newpage
%\fi



%\bibliographystyle{IEEEtran}
%\begin{thebibliography}{1}

%  \bibitem{IEEEhowto:kopka}
%  H.~Kopka and P.~W. Daly, \emph{A Guide to \LaTeX}, 3rd~ed.\hskip 1em plus

%\end{thebibliography}
\bibliographystyle{IEEEtran}
\bibliography{references}

%  Biography text here.
%\end{IEEEbiography}


%\newpage



%\vfill


\end{document}